%% file: main.tex
\PassOptionsToClass{11pt,twocolumn}{article}
\documentclass{hku_style}
\newif\ifdraft
\draftfalse

\usepackage[utf8]{inputenc} 
\usepackage[T1]{fontenc}
\usepackage{url}
\usepackage{nicefrac}
\usepackage{multicol}
\usepackage{microtype}
\usepackage{geometry}
\usepackage{graphicx}%
\graphicspath{{fig/}{assets/}}
\usepackage{multirow}%
\usepackage{amsmath}
\usepackage{amssymb}
\usepackage{amsfonts}
\usepackage{stmaryrd}
\usepackage{mathrsfs}%
\usepackage{xcolor}%
\usepackage{textcomp}%
\usepackage{manyfoot}%
\usepackage{booktabs}%
\usepackage{algorithm}%
\usepackage{tabularx}%
\usepackage{algorithmicx}%
\usepackage{algpseudocode}%
\usepackage{listings}%
\usepackage[inline]{enumitem}
\usepackage{xspace}
\usepackage{tikz}
\usepackage{makecell}
\usepackage{longtable}
\usepackage{changepage}
\usepackage{color}
\usepackage{colortbl}
\usepackage{pifont}
\usepackage{subcaption}
\usepackage{physics}
\usepackage{diagbox}
\usepackage{wrapfig}
\usepackage{siunitx}
\usepackage[numbers]{natbib}
\usepackage{array}
\usepackage{xltabular}
\usepackage{ltablex}
\usepackage{hyphenat}
\keepXColumns

\usepackage{tcolorbox}
\tcbuselibrary{skins, breakable}

\usepackage{hyperref}
\hypersetup{
  colorlinks,
  linkcolor=hkucolor,
  citecolor=hkucolor,
  urlcolor=hkucolor
}
\usepackage[nameinlink,noabbrev]{cleveref}

\usepackage{setspace}
\usepackage{changepage}
\usepackage{float}
\usepackage{tablefootnote}
\usepackage{threeparttable}
\usepackage{placeins}

\input{tex/math_commands.tex}

\input{tex/macros}

\ifdraft
    \providecommand\todo[1]{[\textcolor{red}{TODO: {#1}}]}
\else
    \providecommand\todo[1]{}
\fi

\setlist{nosep}

\newcolumntype{Y}{>{\raggedright\arraybackslash}X}
\newcommand{\corrauth}{\ensuremath{\dagger}}

\author[]{Siyu Liu$^{1,2,*}$, Bo Hu$^{1,*}$, Beilin Ye$^{1,*}$, He Cao$^{3}$, David J. Srolovitz$^{1,2,\corrauth}$, Tongqi Wen$^{1,2,\corrauth}$}

\affiliation{$^1$ Center for Structural Materials, Department of Mechanical Engineering, The University of Hong Kong, Hong Kong, China}
\affiliation{$^2$ Materials Innovation Institute for Life Sciences and Energy (MILES), HKU-SIRI, Shenzhen, China}
\affiliation{$^3$ International Digital Economy Academy (IDEA), Shenzhen, China}

\contribution[*]{Equal contribution}
\hkudata[Corresponding authors$^{\corrauth}$]{\url{tongqwen@hku.hk}, \url{srol@hku.hk}}
\title{Harnessing agent memory to build lifelong AI partners for materials scientists}

\abstract{
\small{
Materials research advances through accumulated experience --  scripts that work,  protocols that are trusted,  warnings attached to failed calculations or experiments, and  judgement that links a new question to an old result. This experience is essential for reproducibility and knowledge transfer, yet it is usually fragmented across notebooks, repositories, job logs and individual memory, and it is rarely portable across artificial-intelligence agents. Here we argue that a lifelong AI partner for materials science can be designed around persistent memory rather than around a particular agent implementation. We introduce a self-evolving memory framework that stores scientific experience as inspectable facts and executable skills, so that observations, failure boundaries, protocols and validation checks can be retrieved, revised and migrated across models. We evaluate the idea in three computational settings that expose different layers of materials-research competence. In 49 real-world materials-tool-use questions comprising 138 executable subtasks, memory nearly doubles GPT-5.2 task success 
without model-parameter updates. In elemental-solid equation-of-state calculations, memory converts a wavefunction-initialization failure into a pre-execution guardrail, improving outcomes from 22/1/4 to 25/2/0 Correct/Partial/Error and avoiding 92\% of repeated errors. In 13 practical material simulation workflows, remembered skills and failure facts halve the aggregate trace burden (tokens) and reduce tool calls by over a factor of two 
by the third round, while preserving physically meaningful outputs in band-gap, phonon, vacancy and work-function analyses. These results show that agent memory can serve as a durable scientific asset;  a portable, self-improving record of materials-research experience that  outlives any single model or agent stack.
}
}

\begin{document}

\maketitle

\section{Introduction}
\label{sec:introduction}

The most valuable companion to a materials scientist, as for all scientists, is not a single instrument, a single database or a single model; it is accumulated research memory. A scientist learns the field by reading papers, attending seminars and discussing with colleagues. They learn how the work is actually done by running calculations, synthesizing samples, debugging tools, writing scripts, preparing figures and discovering which apparently reasonable paths fail. Over years, these experiences crystallize into scientific taste: the ability to recognize opportunities and/or suspicious results, choose a reliable protocol and link a new question to an old lesson. The cost of forgetting is concrete and recurring. In experimental materials research, noting that a precursor must be dried longer than the protocol states, that nominally identical annealing schedules leave metastable phases unless the furnace is cooled in a particular way, or that a weak spectral feature is caused by surface preparation rather than by a distinct phase can decide whether a research team spends days rediscovering a known trap or moves directly to the scientific question. In computational research, the same pattern appears as reusable judgement about which settings make phase-stability comparisons meaningful, which structures must be relaxed before a vibrational or electronic-property calculation is meaningful, and which recurring failures are the result of numerical rather than physical issues. Such memories are not merely workflow conveniences: they determine whether materials claims are comparable, reproducible and credible. Yet today this memory lives in the mind of the scientist, handwritten or digital notes, in code repositories, in failed-job logs and in analysed rather than raw data. The fragments are precious but discrete, unformatted and densely connected, which makes them difficult to retrieve at the moment of need and prone to decay when research personnel move on, when folders are reorganized, when models change or when the original context is forgotten.

The recent wave of artificial intelligence has dramatically accelerated individual steps of the materials-research loop. Autonomous agents can now plan and execute end-to-end chemical synthesis~\citep{Boiko2023Coscientist,Bran2024ChemCrow,White2025PromptCS}, mobile robotic materials scientists can drive multi-instrument self-driving laboratories such as A-Lab~\citep{Burger2020MobileChemist,Dai2024MobileChemist,Szymanski2023ALab}, and multi-agent stacks can coordinate literature reading, experimental design and hardware execution on demand~\citep{Ruan2024LLMRDF,Song2025ChemAgents}. Predictive and generative models have expanded the known space of stable phases (by an order of magnitude), unlocked controllable inverse design, and been transferred to alloy, macromolecule and inorganic-compound generation~\citep{Merchant2023GNoME,Zeni2025MatterGen,Park2025AlloyGPT,Schmidt2024MacromoleculesLLM,Kim2024InorgSynthLLM,Lee2025SAMR}. On the computational side, multi-agent frameworks weave LLM reasoning, knowledge graphs and physics-aware simulations for applications in alloy design, protein discovery, metal--organic framework prediction, organic-semiconductor optimization and broader multi-task computational materials science~\citep{Ghafarollahi2025AtomAgents,Ghafarollahi2025SciAgents,Ghafarollahi2024ProtAgents,Kang2024ChatMOF,Kang2025MOFData,Zhang2024OrgSemiAgent,Chaudhari2025MatSciAgent}, while DFT- and interatomic-potential-oriented agents automate large parts of atomic-scale workflows, and tool-augmented agents now operate user facilities, atomic force microscopes and simulation pipelines under human-in-the-loop supervision~\citep{Wang2025DREAMS,Hu2026TritonDFT,Mathur2024FacilityTools,Mandal2025AFMAgent,Ye2025AMLP,Zhang2025ASA,Wang2025SLMMatrix,Ansari2024Eunomia}. The paradigm has even been pushed to fully automated paper writing~\citep{Lu2026AIScientist}, and reviews are starting to characterize an emerging ``generalist material intelligence'' across these threads~\citep{Yuan2025GMI,Ramos2025ChemAgentReview,Madika2025AINano,Behler2025AICatalysis}. Together, these advances make individual scientific actions measurably faster and more autonomous. Yet each effort is built around the agent rather than around what the agent has learned, so the operational knowledge produced inside one system rarely survives the next model release or framework change. The closest attempts to address this are systems that move toward skill acquisition or tool evolution: CASCADE consolidates memory through continuous learning and self-reflection~\citep{Huang2026CASCADE} and test-time tool evolution synthesizes executable tools as inference-time artifacts~\citep{Lu2026TTE}. Yet even these treat the agent as the carrier of progress; the accumulated experience does not exist as an inspectable, portable object.

The open problem is therefore not whether agents can act faster or accumulate tools inside one framework, but whether the \emph{experience} of the scientist---operational knowledge that carries provenance, boundary conditions and failure history---can persist across models, projects and the inevitable framework evolution. The distinction is important: a tool or skill can encode how to perform a procedure, whereas research experience also includes the conditions under which a procedure is trustworthy, which failure produced a warning, what evidence supports the validity of a boundary condition and whether another model or project should inherit it. Three structural problems converge. First, the agent is treated as the unit of progress: when a stronger foundation model arrives, or when the orchestration framework changes, the operational knowledge accumulated inside the previous agent---what works with a given machine learning potential, where a given DFT functional fails, which k-point sampling is a sufficient anchor for a class of compounds---does not automatically migrate, and there is no consensus on how to even quantify ``what the system has learned'' across runs of a self-driving lab~\citep{Volk2024SDLMetrics}. Second, large language models hallucinate confidently and inconsistently~\citep{Farquhar2024Hallucination,Chen2024PromptHall}, so a one-off conversation history cannot be trusted as a knowledge source without explicit verification, and recent benchmarks of LLM agents on real laboratory instruments show that even strong models fail to translate domain question-answering ability into reliable execution~\citep{Mandal2025AFMAgent}. Third, when the model itself is updated, naive continual learning causes catastrophic forgetting of previously acquired domain skills~\citep{Wang2024CLSurvey}. Existing agent-side memories address only fragments of this need: chains of thought retained inside a single trajectory~\citep{Yao2023ReAct}, verbal post-mortems written after a failed attempt~\citep{Shinn2023Reflexion}, environment-specific code skills indexed by embedding~\citep{Wang2023Voyager}, OS-style virtual context paging~\citep{Packer2023MemGPT}, or episodic logs designed for plausible behaviour in a sandbox~\citep{Park2023GenerativeAgents}. None of these treat scientific memory as a first-class artifact: human-readable, peer-editable, provenance-linked, model-agnostic and portable across the agent stacks that come and go.

We therefore reframe the issue. The lasting contribution of these tools for a scientist is not the agents, but the memory that outlives the agents and the nature of the research workflow that is sufficiently rich to take advantage of this memory (rather than a, for example, flat conversation log). The core object in our framework is a self-evolving knowledge base composed of two complementary, textual artifacts. \emph{Facts} are stored scientific observations, warnings, interpretations and boundary conditions; e.g., a verified machine learning potential for particular applications, a documented convergence failure, or a calibrated reference value. \emph{Skills} are stored, reusable procedures, scripts, protocols and checklists; e.g., a relax-then-DFPT (density functional perturbation theory) workflow, an EOS-fit script, or a slab work-function pipeline. Both are human-readable and provenance-linked; e.g., a scientist can inspect what the agent saved, edit it, version it across projects, and migrate it to a stronger model when one becomes available. This positions memory itself as a long-lived scientific asset (closer to a laboratory protocol than to a model weight) and the agent as an interface that reads, executes and updates that asset under sandbox-grounded feedback.

\begin{figure*}[!t]
    \centering
    \includegraphics[height=0.75\textheight,keepaspectratio]{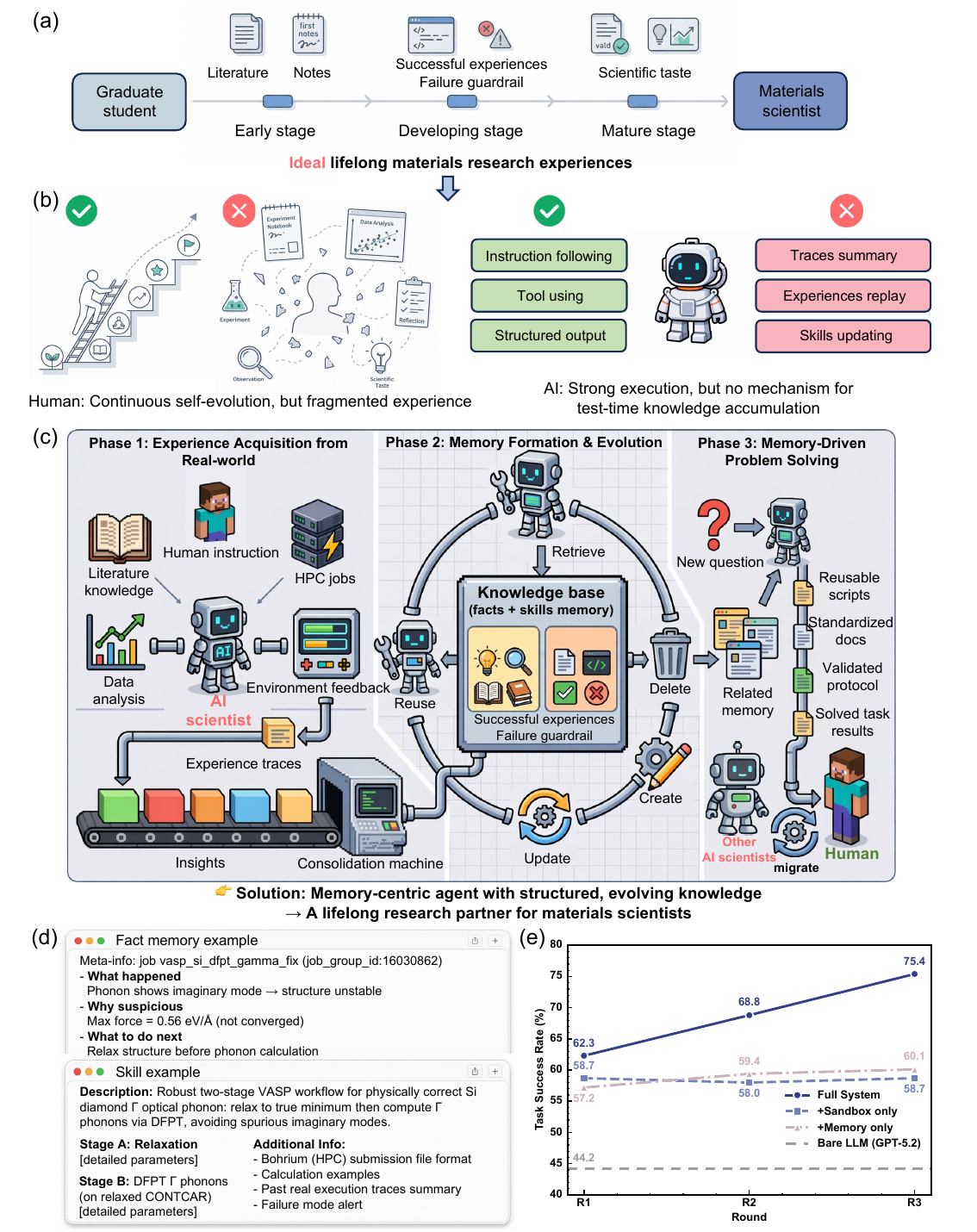}
    \caption{\textbf{Memory-centric self-evolving agent for lifelong materials research.}
(a) Materials scientists develop judgement by accumulating literature, notes, successful experiences, failure guardrails and scientific taste.
(b) Human experience is cumulative but fragmented, whereas current AI agents execute strongly but usually lack a mechanism for test-time knowledge accumulation.
(c) A memory-centric agent acquires experience from real tasks, consolidates it into structured facts and skills, and retrieves it for new problems.
(d) Examples of fact-memory and skill-memory from a VASP phonon workflow, including job provenance, failure diagnosis, recommended action, DFT relaxation followed by a DFPT phonon calculation and HPC execution notes.
(e) Three-round task-success trends on the MatTools materials-tool-use benchmark show that full memory evolution improves GPT-5.2 from 62.3\% to 75.4\%, beyond sandbox-only and memory-only variants.
}
    \label{fig:system_architecture}
\end{figure*}

\Figref{fig:system_architecture} places this design in a single diagram; i.e., an agent acquires experiences from real tasks, consolidates them into human-readable facts and skills, and retrieves or updates them when a new problem appears, with execution-feedback closing the loop. \Figref{fig:system_architecture}d shows the resulting representation on a concrete VASP phonon failure -- the observed imaginary mode, the maximum force of 0.56~eV/\AA{} and the recommendation to relax before DFPT are stored as a fact with job provenance, while the corresponding skill records DFT relaxation followed by a Gamma-point DFPT phonon calculation, together with HPC submission details and failure alerts.

We instantiate this design as a memory-centric agent for computational materials research and probe three layers of materials competence: executable tool use, atomistic-simulation reliability and practical workflow reuse. For example, in the real-world tool-use subset of MatTools~\citep{MatTools} (49 \texttt{pymatgen.analysis.defects} questions, 138 evaluated subtasks), memory raises the GPT-5.2 task success rate from 44.2\% to 75.4\% over three rounds without changing any model parameters, and a memory produced by GPT-5.4 transfers to a weaker GPT-5.4-nano student with a 50.8\% gain over the student model's own three-round memory. On Sol27LC~\citep{Sol27LC}, a recurring wavefunction-initialization failure in ABACUS (DFT) equation-of-state fitting is captured once and avoided in 91.7\% of subsequent cases across structurally related families. In 13 VASP (DFT) and LAMMPS (molecular dynamics) workflows covering band-gap, phonon, vacancy and work-function calculations, retrieved facts and skills halve aggregate token use by the third round while preserving physically meaningful outputs. Together these settings demonstrate that memory is not a benchmark trick but a durable carrier of executable materials-research experience across tasks, sessions and models.

\FloatBarrier

\section{\nohyphens{A memory format for lifelong materials research}}
\label{sec:memory_experience}

A lifelong partner for a materials scientist must remember more than documents; it should remember concepts, decisions, partial attempts, failed boundaries, successful protocols, scripts, parameters and the reasoning that connects them. We use two complementary memory types to span this spectrum. Fact-memory stores compact scientific statements: what happened, in what context, why it matters, what evidence supports it and what should be done next. Skill-memory stores actionable know-how: a goal, applicability conditions, prerequisites,  a procedure, code or parameter notes, validation checks, failure modes and provenance links to the traces that created or revised the skill.

This separation in memory types matters because scientific experience is not uniform. Materials research generates two complementary classes of knowledge that should be remembered in different forms. \emph{Boundary-knowledge} records where a method, parameter or interpretation ceases to be trustworthy: an unrelaxed Si structure that produces an imaginary DFPT phonon, a default wavefunction initialization that prevents SCF convergence, a pseudopotential energy cut-off below which energies drift, or a calibrated lattice constant that should be consulted as a reference rather than reported as a new measurement. \emph{Procedural-knowledge} records the executable know-how that the scientist has gained; e.g., a relax-then-DFPT protocol, an EOS-fit script, a slab work-function pipeline, the sandbox-validated code fragment that turns these protocols into reproducible outputs. Conflating the two is a recurring source of error in agent traces -- a procedure that runs is easily mistaken for a procedure that is trustworthy. We therefore route boundary-knowledge into facts and procedural-knowledge into skills -- facts preserve context, skills preserve ordered procedures, parameters and validation checks, applicability conditions and cautions. Together they enable the system to remember both what to do and what not to do. In particular, a calibrated reference value is stored as boundary-knowledge the agent can consult, not as the answer it should report. This distinction may be seen in the practical-workflow cases below.

A useful representation is one that is intentionally textual and provenance-linked. A text memory is not bound to the weights of one model, the schema of one software stack or the interface of one agent; rather, it can be embedded for retrieval, linked in a graph, inspected by a human, edited, exported to another system or reused by a future model. The same representation evolves with execution feedback; failed runs create guardrails, successful runs create skills, and partially successful runs revise existing procedures after sandbox, job or output validation, so the memory store grows as a curated trace of what has been verified rather than as a passive log of what has been said.

\FloatBarrier

\section{Quantifying memory growth in a controlled test bed}
\label{sec:mattools_benchmark}

Can memory improve executable materials-tool use without updating model parameters? This is the first capability a practical materials agent must acquire -- not simply answering questions about materials science, but selecting APIs, writing code, satisfying structured output contracts and repairing errors after execution. We therefore use the real-world tool-use subset of MatTools~\citep{MatTools}: 49 questions from the \texttt{pymatgen.analysis.defects} test suite, decomposed into 138 evaluated subtasks covering defect construction, vacancy, interstitial and substitution generators, supercell matching, charge-density and local-extrema analysis, formation-energy diagrams, electrostatic corrections, defect-state localization, and radiative or Shockley--Read--Hall recombination calculations -- the full question-level explanation is provided in Supplementary Table 1. This subset is well suited to memory evaluation because many failures cannot be repaired through correcting scientific vocabulary alone; they involve API selection, output schemas, numerical conventions and code that must actually run.

\Figref{fig:mattools_benchmark_result}a compares bare-model performance across several frontier and smaller language models. The three bars separate question pass rate over the 49 top-level questions, task success rate over the 138 evaluated subtasks, and executable-function reliability, which range from 20.4--53.1\%, 23.2--66.7\% and 40.8--89.8\% across the bare-model cohort, respectively. Larger models generally achieve higher values on all three metrics, but even strong models leave substantial room for execution-level improvement: GPT-5.4 achieves an 89.8\% function runnable rate, yet its low task-success and question-pass rates, 66.7\% and 53.1\%, indicate that runnable code alone does not guarantee a correct scientific answer. We then compare each bare model with the full memory-centric system across three rounds. A round is one complete pass over the same 49-question tool-use set -- R1 is a cold-start full-system pass and R2/R3 can retrieve memory accumulated from earlier passes -- no model parameters were trained between rounds. Gains are largest where the baseline has enough competence to produce useful traces but still makes repeated tool-use mistakes. \Figref{fig:mattools_benchmark_result}b shows that GPT-5.2 improves from a bare task-success rate of 44.2\% to 75.4\% after three rounds, GPT-5.4 improves from 66.7\% to 88.4\%, GPT-5.4-mini rises from 39.1\% to 49.3\%, GPT-5.4-nano from 31.2\% to 33.3\% and Qwen3.5-397B from 26.1\% to 33.3\%. Smaller models therefore benefit, but less reliably when they cannot operationalize the retrieved memory.

The ablations in \Figref{fig:system_architecture}e clarify why the full system matters. Sandbox-only execution improves robustness by catching errors but rarely preserves the correction for later sessions and memory-only operation can recall prior notes but without execution feedback risks retaining incomplete or incorrect procedures. The full system couples both -- sandbox signals decide what should be trusted and memory carries the trusted correction forward. The result is a gradual rise from 62.3\% to 75.4\% for GPT-5.2 over three rounds, while sandbox-only and memory-only variants stay between 57\% and 60\%.

Memory also changes the economics of tool use. \Figref{fig:mattools_benchmark_result}c plots task-success improvement against (additional) median token-use relative to bare models, with dashed reference lines reporting efficiency in percentage points of task-success gain per 1{,}000 additional median tokens. This improvement is not free -- retrieval, validation and memory updates add context. GPT-5.4 and GPT-5.2 occupy the high-gain region but at different token costs: GPT-5.4 gains 21.0--21.7 percentage points in R2--R3 with about \(1.3\times10^4\) additional median tokens per question, whereas GPT-5.2 gains 24.6--31.2 points with only \(4\times10^3\)--\(5\times10^3\) additional tokens. Thus, although GPT-5.4 achieves a higher final task-success rate than GPT-5.2 (88.4\% versus 75.4\% in R3), GPT-5.2 uses memory substantially more efficiently, achieving 6.05--6.45 percentage points of task-success improvement per 1{,}000 additional tokens, compared with 1.59--1.69 for GPT-5.4. Smaller models show weaker or even negative movement in some rounds, indicating that extra memory simply becomes overhead when the target model cannot operationalize it and that the relevant axis for choosing a memory-enabled configuration is gain per added token rather than gain alone.

\FloatBarrier

\section{\nohyphens{Memory can migrate between models}}
\label{sec:transfer}

A lifelong research memory should outlive any particular agent. This is especially urgent today, because agent architectures and foundation models change quickly; a system designed around one model may be obsolete when the next model is released, while a well-written protocol, warning or scientific interpretation can remain useful for years. We tested this portability by allowing one model to use memory generated by another.

\begin{figure*}[!t]
    \centering
    \includegraphics[width=0.86\textwidth]{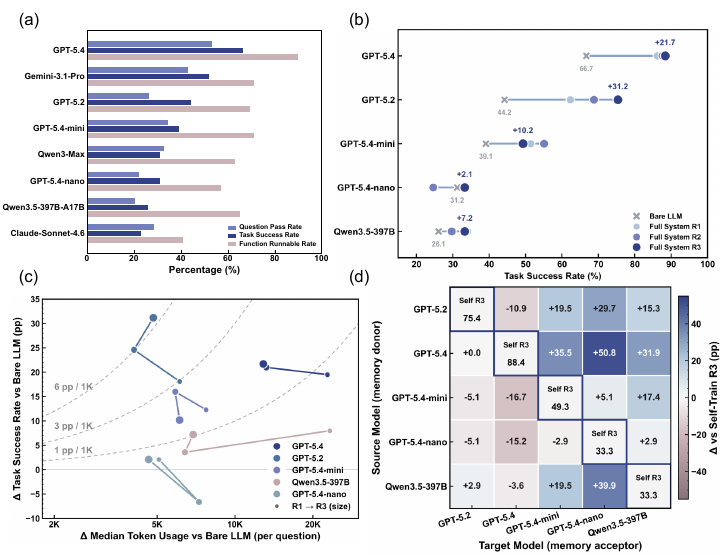}
    \caption{
\textbf{MatTools benchmark performance and memory-driven improvements across LLMs.}
(a) Comparison of question pass rate over 49 top-level questions, task success rate over 138 evaluated subtasks, and function runnable rate across models on the real-world tool-use subset of MatTools.
(b) Task-success rate improvement from bare LLM to the full system across R1--R3. R1 is a cold-start full-system pass; R2 and R3 reuse memory accumulated from previous passes, with model parameters fixed throughout.
(c) Trade-off between task-success gain and additional median token use relative to bare LLM execution; larger markers indicate later rounds. Absolute token increments are shown because they quantify the additional computational cost directly and support the efficiency contours expressed as percentage-point gains per 1{,}000 additional tokens.
(d) Cross-model memory transfer. Off-diagonal cells report the change in target-model task-success rate when using memory produced by a source model after its own three-round run, relative to the target model's R3 memory; diagonal cells show each model's R3 task-success rate.
}
    \label{fig:mattools_benchmark_result}
\end{figure*}

The resulting transfer matrix in \Figref{fig:mattools_benchmark_result}d is asymmetric, as expected. Memories from stronger source models often help weaker ones more than memories from weaker sources help stronger targets. For example, GPT-5.4 memory raises GPT-5.4-nano performance by 50.8 percentage points over the nano model's R3 memory and improves GPT-5.4-mini by 35.5 percentage points. Conversely, memories from smaller models can be neutral or harmful for stronger targets, because the stored procedures encode narrower or less reliable reasoning.

This portability is  important in practice because materials laboratories rarely have uniform access to the same compute, model or expertise. It also resembles knowledge transfer within a research group; validated operational know-how survives personnel turnover only when it is written down as an inspectable protocol rather than left in the mind of a single individual. A validated skill produced during an expensive run with a stronger model can therefore be reused by a cheaper or smaller model when the skill is written in executable, inspectable form. The memory object behaves more like a scientific protocol than a hidden model weight; it can be read, checked, edited and migrated across models, architectures and systems. The same approach also exposes a limitation. Cross-model transfer is not automatically beneficial; several cells are near zero or negative when the source memory is less reliable than the target's own experience. Portability must therefore be paired with provenance and validation rather than treated as blind imitation.

\FloatBarrier

\section{Mechanisms behind memory gains}
\label{sec:mechanism_cases}

\begin{figure*}[!t]
\centering
\includegraphics[width=\textwidth]{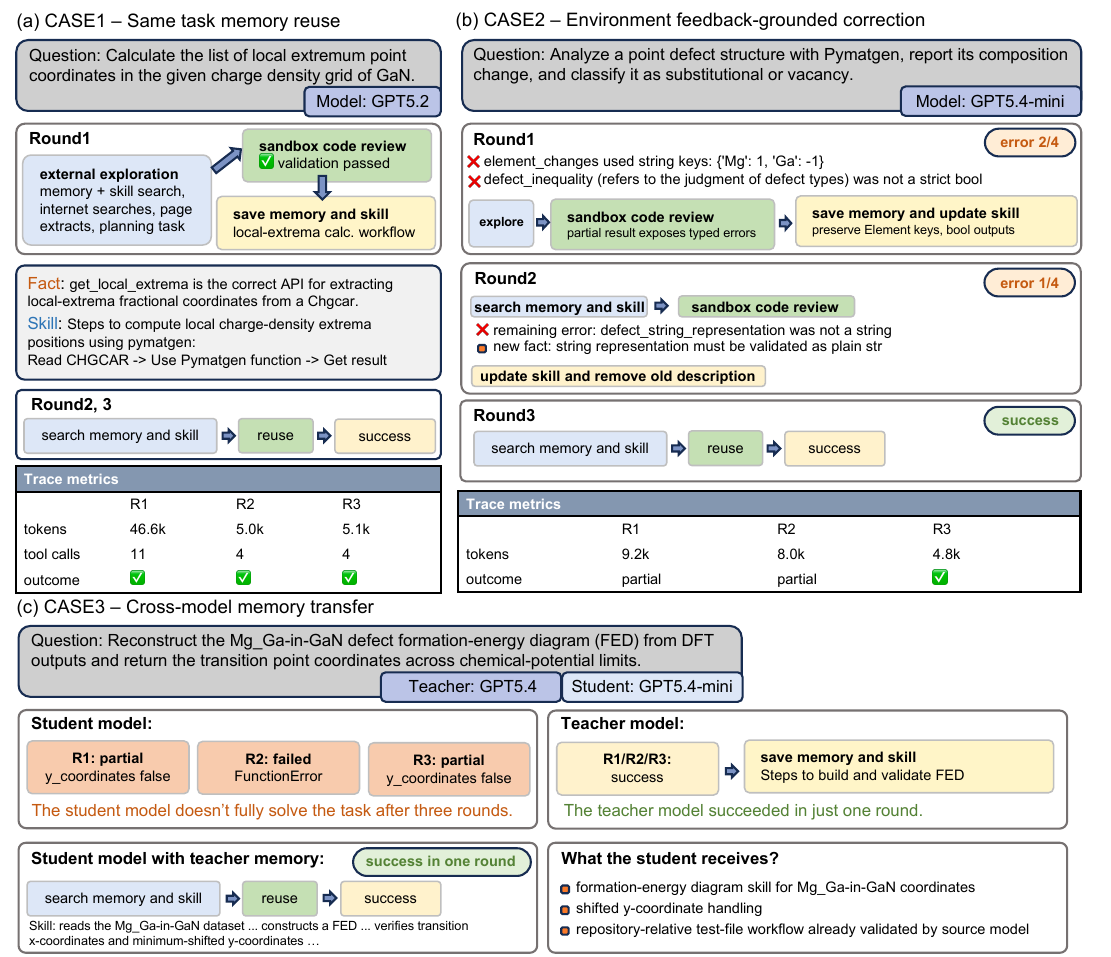}
\caption{
\textbf{Execution-level mechanisms behind memory gains.}
(a) Same-task memory reuse turns an exploratory first run into short later executions by retrieving saved API facts and workflow skills.
(b) Feedback-grounded correction converts sandbox errors into memory facts and skill updates, progressively repairing concrete implementation mistakes.
(c) Cross-model memory transfer lets a student model solve a previously unsolved formation-energy diagram task by reusing a teacher model's validated skill and supporting facts.
}
\label{fig:mattools_case}
\end{figure*}

The aggregate gains are easiest to read through execution traces. \Figref{fig:mattools_case} shows three mechanisms by which memory changes the lifecycle of a scientific task; direct same-task reuse, feedback-grounded repair and teacher-to-student transfer.

In the first case, the task asks for local extrema in a GaN charge-density grid. The first round is exploratory; the system searches memory and skills, consults external sources, extracts pages, plans the task and validates the code. A useful result is not only an answer but the memory distilled from the attempt; \texttt{get\_local\_extrema} is an API for extracting fractional coordinates from a \texttt{CHGCAR}, and the workflow should read \texttt{CHGCAR}, call the Pymatgen function and return the extrema. In later rounds this memory turns an exploratory task into a short reusable procedure; 46.6k tokens and 11 tool calls in R1 collapse to about 5k tokens and 4 tool calls in R2 and R3.

In the second case, the task is to analyse a point-defect structure with Pymatgen. The first round is only partially correct; sandbox review exposes textual-output errors, including string keys in \texttt{element\_changes} and non-boolean defect classifiers. Instead of treating this as a disposable failure, the system saves a correction and updates the skill. The second round exposes one remaining schema problem, which is again consolidated. By the third round, the corrected memory succeeds. This case illustrates why a lifelong memory must store failures as well as successful protocols.

The third case probes whether a memory can act as a transferable scientific object. A GPT-5.4-mini student fails to fully reconstruct a formation-energy diagram for Mg\_Ga-in-GaN substitutional point defects even  after three self-rounds. A GPT-5.4 teacher succeeds and saves a skill describing  formation-energy diagram construction, the shifted y-coordinate handling and a repository-relative validation workflow. When the student uses the teacher memory, it solves the task in one round. The improvement is not bound to the teacher model itself; it is carried by a validated textual procedure that the student can retrieve and execute.

\section{Preventing repeated failures}
\label{sec:sol27lc}

Can memory prevent repeated failures in a workflow?  Consider the case of a standard atomistic-simulation workflow:  i.e., the calculation of the equation-of-state from lattice-constant fitting from DFT calculations (this is known to be very sensitive to input preparation, pseudopotential compatibility, unit conventions, SCF convergence and fit validation). We evaluated this framework on Sol27LC, a benchmark of 27 cubic elemental solids with experimental lattice constants spanning face-centered cubic (FCC), body-centered cubic (BCC) and diamond crystal lattices~\citep{Sol27LC}. Each case computes an equilibrium lattice constant through DFT equation-of-state fitting performed with ABACUS~\citep{Zhou2025ABACUS}, an open-source DFT software package and less familiar to general-purpose language models than common code-analysis libraries. Cases with the same crystal structure share a memory system. The first case in each family is a cold start (FCC Cu, BCC Li and diamond C), and later cases reuse memories and skills accumulated within that family. The experiment therefore isolates whether a memory system can preserve a calculation failure as an actionable guardrail and prevent its recurrence on chemically distinct yet structurally related materials.

\begin{figure}[H]
    \centering
    \includegraphics[width=\columnwidth]{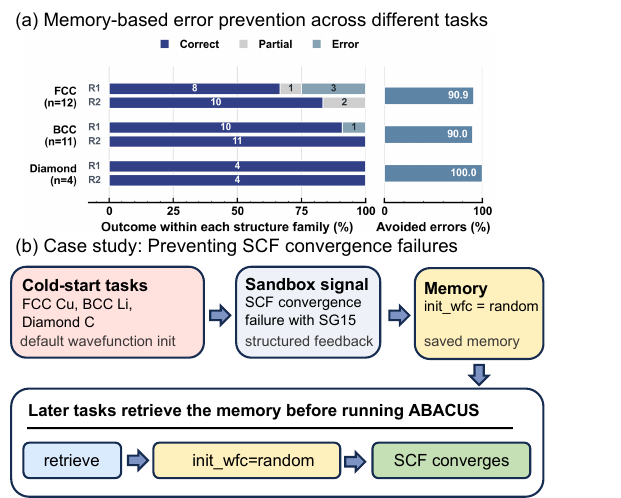}
    \caption{
    \textbf{Sol27LC evaluation and memory-based prevention of repeated convergence failures.}
    (a) Across 27 Sol27LC elemental-solid EOS-fitting calculations grouped by crystal structure, Round 2 reruns eliminate all failed cases, improving outcomes from 22/1/4 to 25/2/0 Correct/Partial/Error and achieving an overall avoided-error rate of 91.7\%. Partial denotes a valid EOS calculation whose final report carries a Bohr-to-\AA{} unit inconsistency. The avoided-error rate measures the fraction of cold-start failures that are successfully prevented in subsequent runs through the retrieval of memories distilled from previous failures.
    (b) Case-study trace showing how cold-start SCF convergence failures with default wavefunction initialization are converted into a reusable memory, \texttt{init\_wfc=random}, which is retrieved by later cases before running ABACUS and prevents repeated convergence failures.
    }
    \label{fig:sol27lc_memory_prevention}
\end{figure}

\Figref{fig:sol27lc_memory_prevention}a shows the result. In the first round, the 27 cases yield 22 Correct, 1 Partial and 4 Error outcomes. Correct means the fitted lattice constant deviates by less than 5\% from experiment. Partial denotes a scientifically valid EOS calculation with the correct physical result, but with the final lattice constant reported in the wrong unit (ABACUS uses Bohr-units internally, whereas the experimental references are in \AA). Error denotes failed runs or large deviations. After Round 2 reruns with accumulated memory, all Error cases are removed and the outcome distribution improves to 25 Correct, 2 Partial and 0 Error. Within the crystal lattice structure families, FCC improves from 8/1/3 to 10/2/0 Correct/Partial/Error, BCC improves from 10/0/1 to 11/0/0, and diamond remains 4/0/0. 

The case study in \Figref{fig:sol27lc_memory_prevention}b explains how this aggregate improvement arises. In the cold-start cases, self-consistent field (SCF) calculations with SG15 ONCV pseudopotentials \citep{Schlipf2015SG15} can fail under the default wavefunction initialization. The useful correction is simple but operationally specific; i.e., set \texttt{init\_wfc=random} before running ABACUS. Once this intervention is distilled into memory as a reusable skill, later cases retrieve it before execution and avoid the same convergence failure in 90.9\% of the FCC cases, 90.0\% of the BCC cases and 100.0\% of the diamond cases (right panel in \Figref{fig:sol27lc_memory_prevention}a). The overall avoided-error rate is 91.7\%, measuring whether cases avoid the repeated wavefunction-initialization failure rather than whether every final report is free of formatting or unit mistakes. The unit of generalization is the crystal structural family: a fix discovered on a single cold-start element propagates as a durable pre-execution guardrail to every chemically distinct member of the same family, so that an intervention validated on one FCC, BCC or diamond element protects subsequent calculations across that family without further human input. This shows that memory captured at the right level of abstraction generalizes across compositions rather than being tied to the specific material on which it was first observed, and mirrors the function of effective laboratory memory; i.e., a local failure is converted into a structural-family-level pre-execution warning that protects later work without external intervention.

\section{Practical computational  workflows}
\label{sec:real_tasks}

Can the same memory reduce repeated setup and analysis costs in practical workflows rather than only in controlled benchmarks? We  evaluated the system on 13 computational materials tasks that resemble routine research work in a research group. The tasks include VASP (a DFT package) calculations of band structures, phonons, dielectric constants, effective masses, surfaces and work functions, as well as LAMMPS (a molecular dynamics package) calculations for vacancy and thermal properties; the task IDs and references are listed in Supplementary Table 2. Although lifelong research memory extends beyond computation, these tasks provide a measurable analogue of the broader materials-research loop; i.e.,  defining a question, gathering knowledge, preparing a protocol, executing it, checking the result, revising the workflow and preserving what was learned.

\begin{figure*}[!t]
    \centering
    \includegraphics[width=\textwidth]{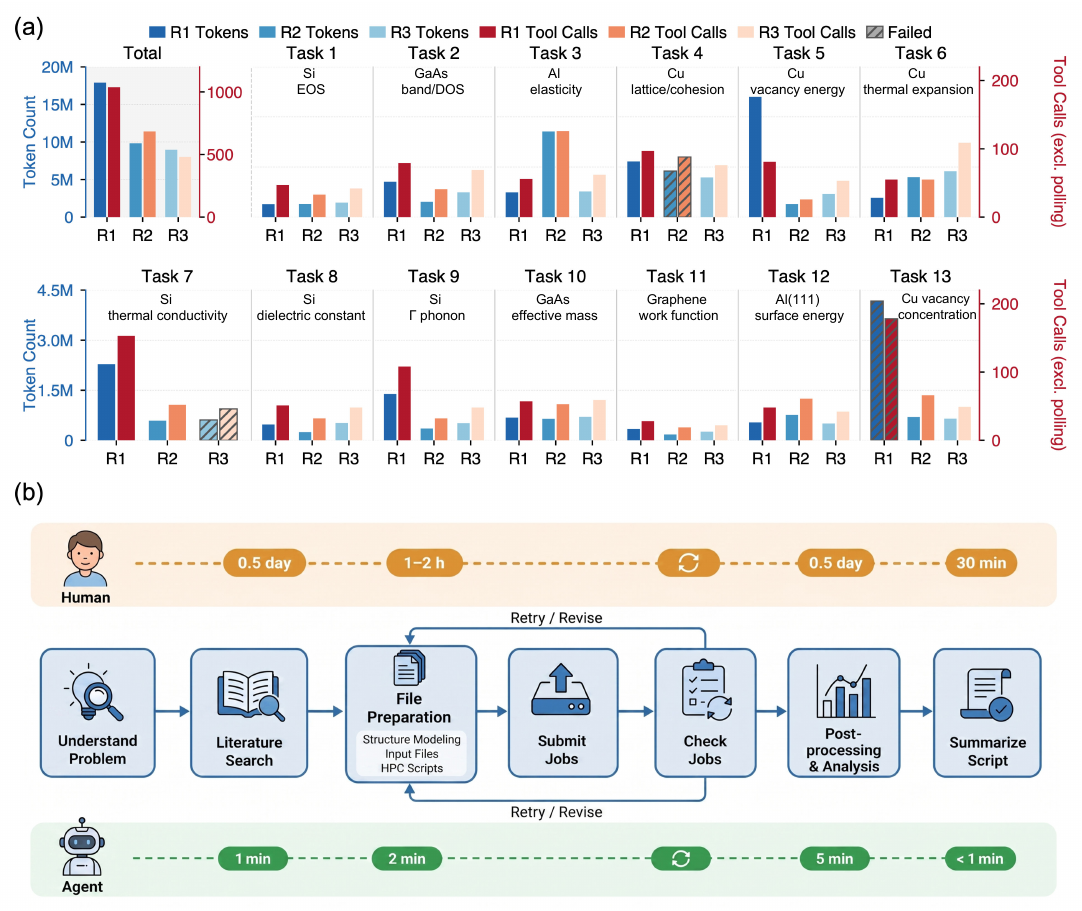}
    \caption{\textbf{Benchmarking performance on practical computational materials science tasks.}
(a) Token consumption and non-polling tool calls across three rounds for 13 VASP and LAMMPS tasks. Total tokens decrease from 17.90M to 9.84M and 8.96M over R1--R3, while tool calls decrease from 1{,}038 to 684 and 481. The aggregate trace burden decreases after R1, but failures and extra checks remain task-specific and non-monotonic. Hatched bars indicate rounds that did not yield a valid final result; their heights still report the tokens and tool calls consumed during those failed rounds.
(b) Human-to-agent time-scale comparison for a recurring computational materials science workflow. Human execution typically requires hours to days for problem framing, literature search, file preparation, monitoring, post-processing and summarization; after memory accumulation, the agent can reuse validated prior traces to perform the repeated preparation and analysis steps on minute-scale time budgets. The comparison highlights superhuman efficiency in routine workflow reuse, while job execution and scientific validation remain part of the process.
}
    \label{fig:real_task_result}
\end{figure*}

The central efficiency gain is best understood as a compression of repeated cognitive and clerical work. For a human researcher, the recurring parts of a familiar computational workflow commonly unfold on hour-to-day time scales; i.e., framing the problem, locating prior scripts, preparing input files, checking convergence parameters, monitoring jobs and post-processing results. After memory accumulation, the agent can recover problem context, prior scripts, input templates and parameter choices on minute-scale time budgets, then reuse them during file preparation, analysis and summarization. \Figref{fig:real_task_result}b places this contrast on a concrete time axis.

\Figref{fig:real_task_result}a quantifies the same reduction across 13 tasks. Total token use decreases from 17.90M in R1 to 9.84M in R2 and 8.96M in R3, while non-polling tool calls decrease from 1{,}038 to 684 and then 481, reaching a 50.0\% token reduction and 53.7\% tool-call reduction by R3. The largest drops occur where prior traces become directly reusable assets; i.e., the Cu monovacancy formation energy calculation falls from 3.60M to 387.3k tokens in R2, the Si thermal conductivity from 2.28M to 586.1k, the Si $\Gamma$-point optical phonon from 1.39M to 350.9k, and the Cu equilibrium vacancy concentration from 4.17M to 700.2k after its initial failed run.

The per-task data also show why memory should be interpreted as workflow reuse rather than automatic compression. Some tasks become heavier in later rounds; e.g., the fcc Al elastic constant calculations expand to 2.56M tokens and 126 tool calls in R2, and Cu thermal expansion calculations grow in both R2 and R3 as the agent performs additional checks. Individual failures remain non-monotonic; e.g., the Cu lattice-constant and cohesive-energy task (Task 4) fails in R2 after a retry-heavy trace, the Si thermal-conductivity task (Task 7) fails in R3, and the Cu equilibrium-vacancy-concentration task (Task 13) fails in R1 despite substantial tool use; overall success is 10/13 in R1, 11/13 in R2 and 9/13 in R3. Memory reduces avoidable rediscovery, but it does not eliminate physical judgement, job variability or the need to verify the current calculation.

\begin{figure*}[!t]
    \centering
    \includegraphics[width=\textwidth]{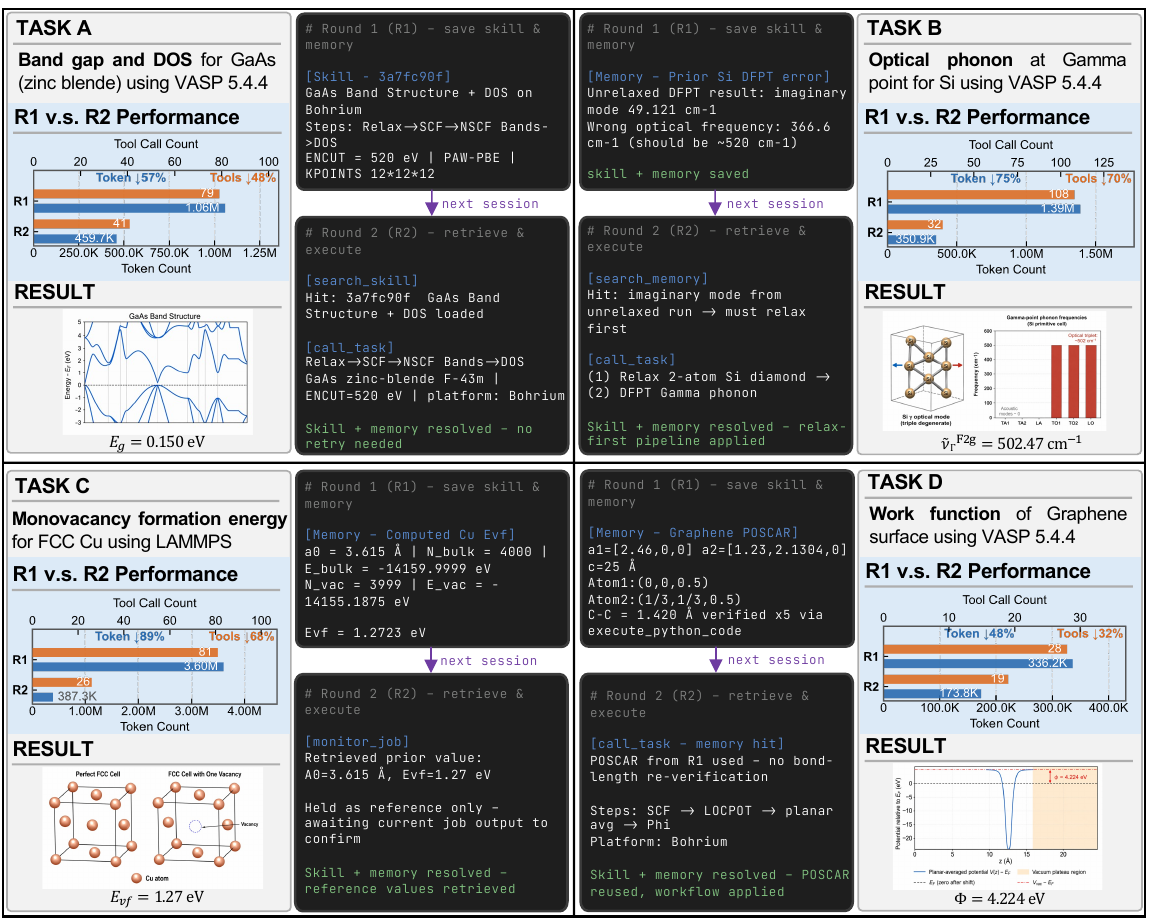}
    \caption{\textbf{Cases showing the effects of memory and skill in practical computational materials science tasks.}
Task A reuses a saved VASP workflow for GaAs band structure and DOS, reducing tokens from 1.06M to 459.7k and tools from 79 to 41 while producing \(E_g = 0.150\) eV.
Task B converts a prior Si DFPT phonon failure into a relax-first memory, reducing tokens from 1.39M to 350.9k and tools from 108 to 32 while yielding \(\tilde{\nu}_{\Gamma}^{F2g}=502.47~\mathrm{cm}^{-1}\).
Task C retrieves prior lattice and formation-energy information for the FCC Cu monovacancy-formation-energy task, reducing tokens from 3.60M to 387.3k and tools from 81 to 26 while preserving the need for current-job confirmation.
Task D reuses a validated graphene POSCAR and work-function workflow, reducing tokens from 336.2k to 173.8k and tools from 28 to 19 while producing \(\Phi=4.224\) eV.
}
    \label{fig:real_task_cases}
\end{figure*}

\Figref{fig:real_task_cases} provides four concrete examples. In a GaAs band-structure and density-of-states task, the first round saves a VASP skill specifying relaxation, SCF, NSCF band and DOS steps with ENCUT (parameter), a \(12\times12\times12\) k-point mesh (parameter) and PAW-PBE (density functional choice). The second round retrieves the skill and produces a band gap of 0.150~eV, consistent with the well-known semilocal-DFT underestimation of GaAs band gaps and therefore better interpreted as a reproduced PBE-level result than as an absolute reference~\citep{MaterialsProjectGaAsMP2534,Vurgaftman2001BandParameters}, with token use dropping from 1.06M to 459.7k and tool calls from 79 to 41. In a Si optical-phonon task, a prior unrelaxed DFPT run had produced an imaginary mode at 49.121~\(\mathrm{cm}^{-1}\) and a spurious optical frequency of 366.6~\(\mathrm{cm}^{-1}\); the saved memory enforces a relax-first pipeline, yielding a Gamma-point optical phonon of 502.47~\(\mathrm{cm}^{-1}\), softened by about 3\% relative to the experimental Raman value near 520~\(\mathrm{cm}^{-1}\), but still physically reasonable for this workflow class~\citep{Vanco2014RamanSi}, with tokens reduced from 1.39M to 350.9k and tool calls from 108 to 32. In an FCC Cu monovacancy-formation-energy task, the system retrieves prior lattice and vacancy-formation-energy information (\(a_0=3.615~\text{\AA}\), \(E_\mathrm{vf}=1.2723\)~eV) as a reference (consistent with copper vacancy literature in the practical-workflow context), while still awaiting current-job confirmation~\citep{Mishin2001Cu,Hehenkamp1992CuVacancy}, reducing tokens from 3.60M to 387.3k and tool calls from 81 to 26. In a graphene work-function task, the agent reuses a validated POSCAR  (crystal structure and computational cell specifications) file with a 25~\AA{} vacuum spacing and a verified 1.420~\AA{} C--C bond length, avoids redundant bond-length verification and obtains \(\Phi = 4.224\)~eV, consistent with the graphene work-function reference used for this task~\citep{Yan2012GrapheneWorkFunction}, with tokens reduced from 336.2k to 173.8k and tool calls from 28 to 19.

These examples display two complementary functions of memory. Skills accelerated repeatable workflows (when the previous protocol is valid) and facts supply cautions (when a previous result should be treated as a reference, warning or sanity check) rather than copied as a final answer.

\section{Discussion}
\label{sec:discussion}

These results shift the unit of progress from the agent to the memory system in AI-for-science systems. If these systems are designed around a current agent, scientific experience remains tied to a model, prompt stack, tool interface or conversation history. If they are designed around memory, the durable asset is the accumulated record of facts, protocols, warnings and validations, while agents become replaceable interfaces that read, execute and revise that record. This framing is especially important in AI-for-science-based research, where a useful lesson is often operational rather than declarative; e.g., a pseudopotential setting, a unit convention, a relax-before-property protocol or a failure mode that should be checked before submitting another job.

Three computational layers support this shift, each probing a different depth of research competence. The first is computational materials competence; MatTools shows whether the agent can execute the right API, schema and code so that a scientific calculation can actually run, and whether the resulting textual record can transfer to a different model. The second is physical reliability; Sol27LC shows that a single, operationally specific failure can be encoded as a pre-execution guardrail that generalizes at the level of the crystal lattice structure family, so that a fix learned on one element protects chemically distinct members of the same family from falling into the same numerical trap. The third is workflow reuse efficiency; the VASP and LAMMPS tasks show that the same memory format compresses the repeated cognitive and clerical steps of a familiar workflow while leaving physical judgement and current-run verification untouched. Taken together, these layers progress from whether a calculation can run, to whether known physical pitfalls can be avoided, to whether prior work can be reused without losing scientific rigor, and memory contributes at every level without changing model weights.

The boundaries are equally important. Memory quality depends on evidence quality; an unvalidated procedure can propagate errors and cross-model transfer can be harmful when the source memory is weaker than the target's own experience. Practical-workflow traces also show that memory is not a universal token-compression mechanism; some tasks become longer when the agent retrieves broader context or performs additional validation. Sandbox review, job feedback, provenance and human inspection are therefore central rather than optional. Text is portable and scientifically legible, but complex multi-file workflows, pseudopotential choices, convergence parameters and material-class assumptions should eventually be paired with stricter schemas, applicability tags, repeated-success counts, deprecation policies and executable validation tests.

Finally, our evidence is computational. The motivation extends to experimental protocols, synthesis know-how, instrument operation, literature judgement and project-level scientific taste; these settings will require their own validation standards. A durable scientific memory should make future models better and more useful rather than making old experience obsolete; reaching that goal  requires memory snapshots, task harnesses, trace release and peer-editable review practices, not only stronger agents.

\section{Methods}
\label{sec:methods}

\begin{figure*}[!t]
    \centering
    \includegraphics[width=\textwidth]{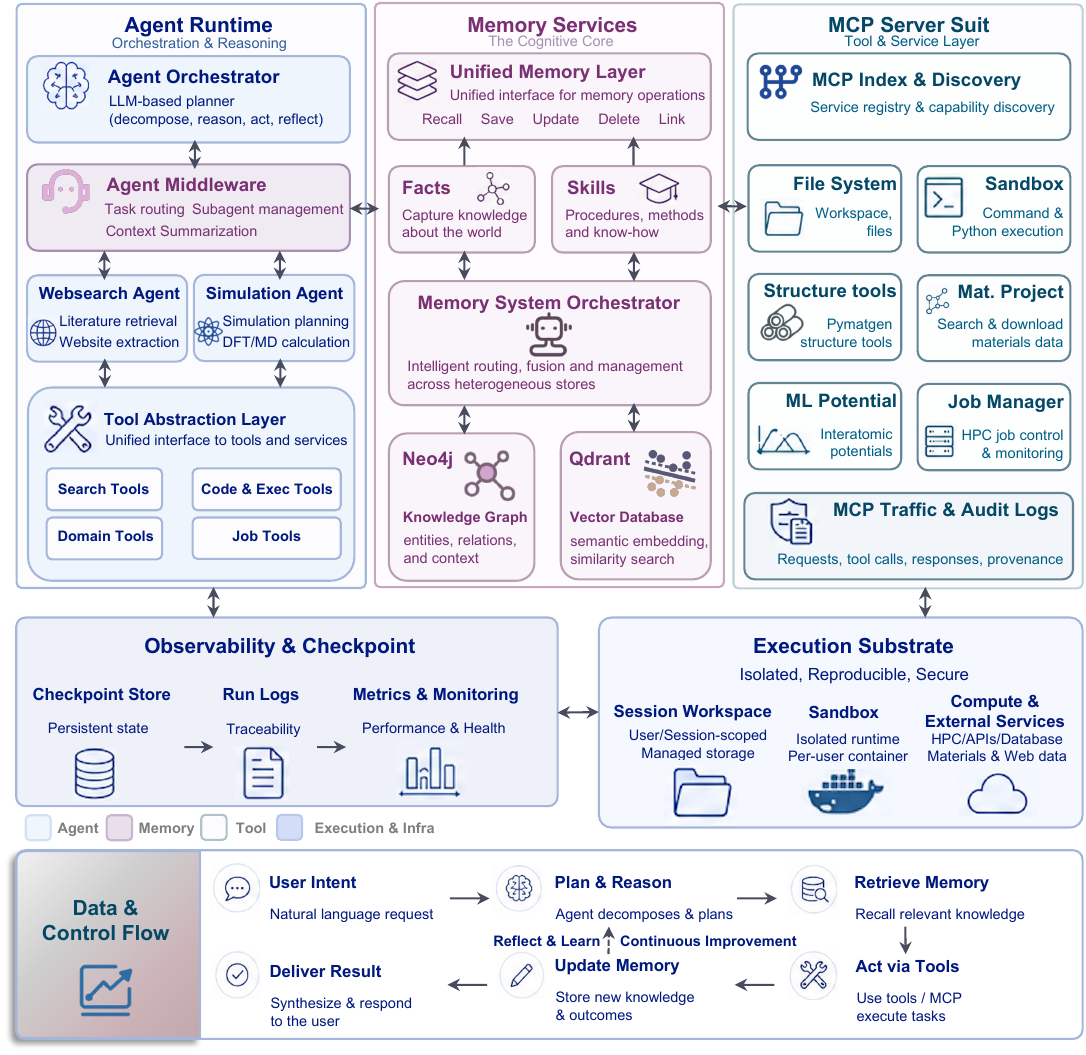}
    \caption{
    \textbf{Overview of the memory-centric agent system architecture.}
    The framework couples an LLM-based agent runtime with unified memory services and an MCP-based tool layer, enabling the agent to decompose user intent, retrieve prior knowledge, execute materials science workflows in sandboxed environments, and update memory from observed outcomes.
    }
    \label{fig:methods_architecture}
\end{figure*}

\subsection{System architecture}

The framework couples a hierarchical agent runtime with a session-scoped tool layer and a long-term memory subsystem (\Figref{fig:methods_architecture}; full interface and prompt definitions in Supplementary Notes 4--6). The runtime is organized around a top-level \texttt{ResearchAgent} that plans the scientific workflow, manages the memory and skill lifecycle, and delegates retrieval to a \texttt{websearch-agent} (literature, URL, PDF, arXiv and DOI-linked queries) and execution to a \texttt{simulation-agent} (sandboxed code and HPC job control). Agent state is checkpointed in a local SQLite database. The tool layer is exposed by an MCP server that mounts FastMCP endpoints behind a FastAPI REST layer served by Uvicorn, with a shared \texttt{session\_id} routing files, sandbox state and job records into per-session workspaces. The MCP layer provides file-system access, sandboxed command and Python execution with a 150s direct-execution timeout, Pymatgen and Materials Project structure tools, NIST interatomic-potential retrieval, and HPC job submission and monitoring; longer simulations are dispatched through the job-management interface and resumed by the agent through an automated monitor prompt. The memory subsystem is built on the open-source mem0 framework with two layers: \emph{Facts} for declarative observations, warnings and parameter choices, and \emph{Skills} for reusable procedures, scripts and protocols. Facts are written with \texttt{infer=True} so that mem0 extracts atomic statements and decides add/update/delete/no-op against existing entries, while Skills are written with \texttt{infer=False} to preserve a complete procedural artefact. Vector retrieval uses Qdrant with 4096-dimensional \texttt{qwen3-embedding-8b} embeddings, and entity--relation memory uses Neo4j with a custom graph-extraction prompt that canonicalizes materials, libraries and methods. Extraction and graph construction use \texttt{qwen3-max} independently of the reasoning model under evaluation, so the memory policy remains stable across model ablations. Memory is scoped by \texttt{user\_id}, with Skills stored under the derived namespace \texttt{user\_id\_skills}.

\subsection{Task and memory lifecycle}

Each task executes a retrieve--plan--act--reflect--update loop. Before expensive actions, the agent calls \texttt{search\_memory} and \texttt{search\_skill} to recover prior facts, scripts and failure warnings. After execution, sandbox or job feedback determines what is consolidated: successful traces register or refine procedures through \texttt{save\_to\_skill} or \texttt{update\_skill}, and new observations, error fixes and parameter choices are written through \texttt{save\_to\_memory}. At login, locally curated \texttt{SKILL.md} files are synchronized against the Skills namespace so that manually authored and learned procedures are retrievable through the same interface. All entries are written in human-readable text and linked to the execution trace that produced them.

\subsection{MatTools benchmark evaluation}

We evaluate on the real-world tool-use subset of MatTools~\citep{MatTools}, comprising 49 questions from the \texttt{pymatgen.analysis.defects} test suite decomposed into 138 evaluated subtasks (full list, evaluated outputs and per-question prompts in Supplementary Note 1). The QA-only portion of MatTools is not used. Question pass rate is the fraction of the 49 top-level questions in which all subtasks pass; task success rate is the fraction of the 138 subtasks that pass; function runnable rate is the fraction of generated functions that execute without runtime, import or output-schema errors in the harness. Five configurations are evaluated and differ only in which subsystems are exposed to the agent and in the corresponding prompt edits: full system, no-memory ablation, no-sandbox ablation, bare-LLM baseline and cross-transfer memory test. Subsystem state per configuration and the exact prompt edits applied to the research, simulation and task-framing prompts are summarised in Supplementary Note 7. Sandbox feedback is exposed only through \texttt{mattools\_code\_check}, which extracts the submitted function, runs it inside an isolated Docker Python environment and returns the captured execution status, \texttt{stdout} and \texttt{stderr}; it never returns the benchmark answer or a corrected implementation. A round is one complete pass over the same 49 questions: R1 is a cold-start pass, and R2--R3 reuse memory accumulated and consolidated in previous passes, with model parameters fixed across rounds. When several reasoning models are evaluated under the full system in the same study, each model receives its own \texttt{user\_id} and memory namespace so that writes do not cross models; the tool semantics, prompts and agent design observed by the model are unchanged. For cross-model transfer, the target model's writes are disabled and the read tools are bound to a frozen source-model namespace produced after that source model's own three full-system rounds; we report the target's task-success change relative to its own R3 memory.

\subsection{Sol27LC evaluation}

Sol27LC contains 27 elemental cubic solids spanning FCC, BCC and diamond structures with experimental lattice constants as references~\citep{Sol27LC}; the structure list and per-case results are tabulated in Supplementary Note 3. Equation-of-state fits are computed with ABACUS~\citep{Zhou2025ABACUS} using SG15 ONCV pseudopotentials~\citep{Schlipf2015SG15}. Initial and follow-up agent prompts used to drive each case are reproduced in Supplementary Note 3. Cases within the same crystal-structure family share a memory store; FCC Cu, BCC Li and diamond C are the cold-start cases. R1 runs the family with memory accumulation enabled, and R2 reruns only the cases that failed in R1, retrieving the accumulated memory before execution. A case is classified \emph{Correct} when the fitted lattice constant deviates from experiment by less than 5\%, \emph{Partial} when the EOS fit reproduces the correct physical equilibrium but the reported value carries a residual Bohr-to-\AA{} unit inconsistency, and \emph{Error} otherwise. The avoided-error rate measures, among R1 failure cases caused by wavefunction-initialization non-convergence, the fraction that no longer fail in R2.

\subsection{Practical computational workflows}

The practical-workflow evaluation comprises 13 VASP and LAMMPS tasks covering equation-of-state, band structure and density of states, elastic constants, cohesive energy, vacancy formation energy, thermal expansion, thermal conductivity, dielectric constant, optical phonons, effective mass, work function, surface energy and equilibrium vacancy concentration; identifiers, descriptions and experimental references are listed in Supplementary Note 2, together with the task-framing and automated-monitor prompts used to drive the agent during long-running jobs. Calculations submitted through the MCP job-management interface return identifiers that are watched by an automated monitor prompt and relayed back to the agent when results are available, so the agent can resume execution and finalize the task. A task is counted as successful only when a physically interpretable final result is delivered; failures cover calculation errors, missing final values and outputs that cannot be validated from the trace. Token counts are computed over the complete agent trace per round, including memory retrieval, code generation, sandbox execution, job submission and monitoring, post-processing and final synthesis. Tool-call counts exclude repeated job-polling calls so that the statistic reflects substantive agent actions rather than scheduler waiting.

\FloatBarrier

\section*{Acknowledgments}
The work described is partially supported by a grant from the NSFC/RGC Joint Research Scheme sponsored by the Research Grants Council of the Hong Kong Special Administrative Region, China and the National Natural Science Foundation of China (Project No. N\_HKU767/25). The authors would also like to thank  the Materials Innovation Institute for Life Sciences and Energy (MILES) for startup funding and HKU-SIRI in Shenzhen for partial support of this work. This work was also partially supported by the Research Grants Council, Hong Kong SAR through the General Research Fund (17210723, 17200424). T. W. acknowledges additional support by the General Research Fund (17211726) and the Guangdong Natural Science Fund (2025A1515012129).

\bibliographystyle{unsrtnat}
\bibliography{main}

\end{document}

%% file: tex/math_commands.tex
\usepackage{amsmath,amsfonts,bm}

\def\Figref#1{Figure~\ref{#1}}

\def\eqref#1{equation~\ref{#1}}

\def\1{\bm{1}}

\DeclareMathAlphabet{\mathsfit}{\encodingdefault}{\sfdefault}{m}{sl}
\SetMathAlphabet{\mathsfit}{bold}{\encodingdefault}{\sfdefault}{bx}{n}



%% file: tex/macros.tex
\definecolor{myorange}{RGB}{220,128,0}
\definecolor{myblue}{RGB}{0,80,220}
\definecolor{myred}{RGB}{220,20,0}
